# Autonomous Slalom Maneuver Based on Expert Drivers' Behavior Using Convolutional Neural Network

Shafagh A. Pashaki
Virtual Reality Lab
Department of Mechanical Engineering, K. N. Toosi University of Technology
Tehran, Iran

Ali Nahvi *
Virtual Reality Lab
Department of Mechanical Engineering, K. N. Toosi University of Technology
Tehran, Iran

Ahmad Ahmadi
Virtual Reality Lab
Department of Mechanical Engineering, K. N. Toosi University of Technology
Tehran, Iran

Sajad Tavakoli
Department of Biotechnology and Biomedicine, Technical University of Denmark
Lyngby, Denmark

Shahin Naeemi
Virtual Reality Lab
Department of Mechanical Engineering, K. N. Toosi University of Technology
Tehran, Iran

Salar H. Shamchi
Virtual Reality Lab
Department of Mechanical Engineering, K. N. Toosi University of Technology
Tehran, Iran

*Corresponding Author: Nahvi@kntu.ac.ir ; Virtual Reality Lab: https://www.drivingsimulator.ir/

*Abstract*— Lane changing and obstacle avoidance are one of the most important tasks in automated cars. To date, many algorithms have been suggested that are generally based on path trajectory or reinforcement learning approaches. Although these methods have been efficient, they are not able to accurately imitate a smooth path traveled by an expert driver. In this paper, a method is presented to mimic drivers' behavior using a convolutional neural network (CNN). First, seven features are extracted from a dataset gathered from four expert drivers in a driving simulator. Then, these features are converted from 1D arrays to 2D arrays and injected into a CNN. The CNN model computes the desired steering wheel angle and sends it to an adaptive PD controller. Finally, the control unit applies proper torque to the steering wheel. Results show that the CNN model can mimic the drivers' behavior with an $R^2$-squared of 0.83. Also, the performance of the presented method was evaluated in the driving simulator for 17 trials, which avoided all traffic cones successfully. In some trials, the presented method performed a smoother maneuver compared to the expert drivers.

*Keywords— lane-changing, obstacle avoidance, convolutional neural networks, deep learning, adaptive PD.*

## I. Introduction

Autonomous driving will be an inextricable part of future automated vehicles. These systems can potentially facilitate the use of cars and reduce accidents. Many car crashes are directly related to human error, fatigue, etc. More than 94% of accidents occur due to driving faults [1]. 539000 accidents occur due to wrong lane-changing in the US annually [2]. So, it is important to equip cars with new smart systems to reduce the number of accidents in lane changes.

To date, various partially and fully autonomous driving systems have been developed including parallel park systems, adaptive cruise control, autonomous emergency braking, and automatic lane-changing system [3]. Among them, automatic lane-changing system is more complex; hence, it is challenging to develop a proper and safe lane-changing system [4]. Therefore, it's essential to precisely develop a lane-changing system, considering the safety of the passengers in addition to avoiding crashes. So, it's significant that the automated car equipped with a lane-changing system must take a safe route smoothly (not aggressively or clumsily) in order to fulfill the passengers' safety and comfort [5].

To overcome the mentioned complexity of the lane-changing system, researchers have developed different methods. Overall, these methods can be divided into two main categories. The first category is path trajectory-based approaches, where a geometric path is designed as the reference path and a controller steers the car on the reference path. The reference path is usually a third-degree or higher-degree polynomial curve, and the controller could be PID, adaptive PID, MPC, fuzzy controller, etc. For example, Wang et al. [6] considered a seventh-degree polynomial curve to develop a lane-changing system. They preferred using a seventh-degree polynomial instead of lower-degree ones as the lower-degree polynomial curves are not as smooth as higher-degree ones [6, 7]. In another work, Chowdhri et al. [8] designed a new automatic system to perform an evasive lane-changing maneuver by tracking a desired reference path leveraging a nonlinear MPC. They also took the brake system into account and entered the dynamics of the brake system in the designed controller.

The second category is reinforcement learning-based approaches through which an agent can learn a task by trial-and-error. These approaches are commonly model-free and make the agent interacts with a stochastic environment based on the states and immediate rewards and the agent tries to maximize the long-term reward. The initial mathematical model of reinforcement learning (RL) algorithms is based on Markov decision process, through which the discrete stochastic environment for the agent is formalized [9]. Aiming at developing an RL-based lane-changing system, Ye et al. [10] devised a new lane-changing system using proximal policy optimization-based deep reinforcement learning. Also, in another endeavor, Mirkevska et al. [11] presented a new reinforcement learning-based approach combined with formal safety verification. They exploited 13 features and a deep Q-network to gain a fast learning rate; the simulation results showed that the designed system worked well.

Although a plethora of works based on the two mentioned categories (path trajectory and RL-based approaches) have been conducted, and these works have achieved great successes, they are not able to imitate the behavior of expert drivers. Path trajectory-based methods are commonly based on a third-degree polynomial which is not perfectly smooth. Even though some papers suggest using higher-degree polynomial curves to circumvent the smoothness problem [6, 7], these paths cannot be a perfect alternative for the path taken by an expert driver. The smoothness problem also exists for RL-based approaches as these methods try to learn the environment based on trial-and-error; therefore, although the agent is able to find its path and finish its mission, the path might not be as smooth as the route taken by a skilled and expert driver.

To address the mentioned gap, we developed a new system based on expert drivers' behavior data. We extracted some features and input them into a convolutional neural network (CNN). The output of the CNN model is the steering wheel angle that is sent to the control unit, which is based on an adaptive PD. The system is designed to mimic expert drivers' behavior. A byproduct of this paper will be to develop a haptic driving training simulation system to train novice drivers. The rest of the paper is organized as follows. The methodology employed in this research is elaborated in section II. The results and the performance of the developed system are presented in section III. Finally, discussion and conclusion are presented in sections IV and V, respectively.

## II. METHODOLOGY

### A. Overview of the developed system

In this research, we devised a new system that is able to mimic expert drivers' behavior for obstacle avoidance and lane-changing. To develop the system, four experienced drivers were asked to perform a slalom maneuver made up of three lane-changing and four longitudinal movement tasks (Fig. 1). Then, the driving data of the drivers were gathered and employed to train a CNN model in order to compute the proper steering wheel angle. Then, the computed steering wheel angle is sent to a control unit which is based on an adaptive PD controller. After that, the control unit calculates an appropriate torque and applies it to the steering wheel. Fig. 2 depicts the steps of the designed system. In the next subsection, we address the data collection process.

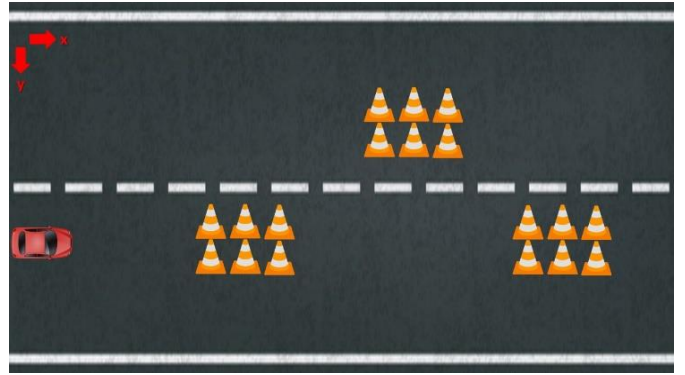

Figure 1. The slalom maneuver used for this research

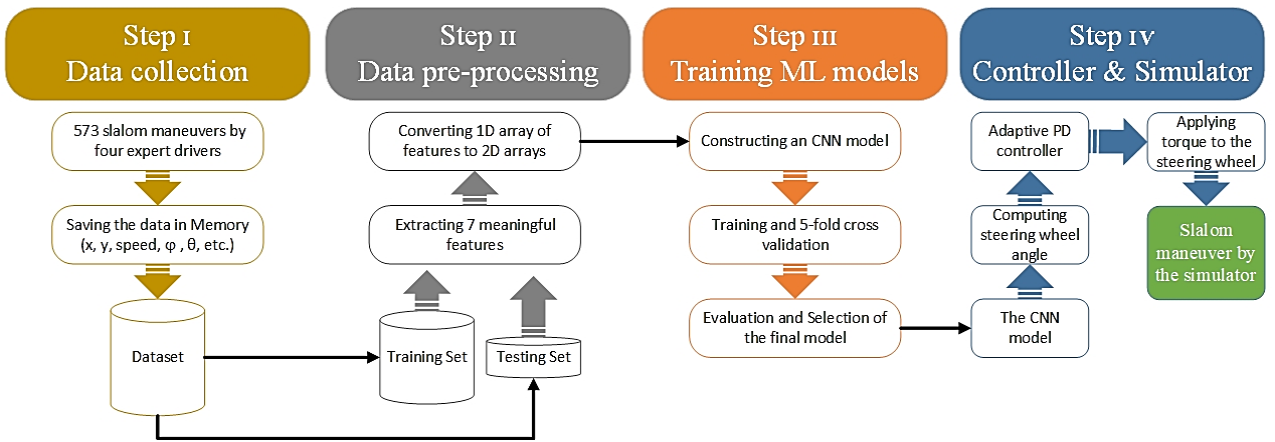

Figure 2. The flowchart of our work. $x$ and $y$ are position of the car. $\varphi$ is the car head angle relative to $y$-axis. $\theta$ is the steering wheel angle.

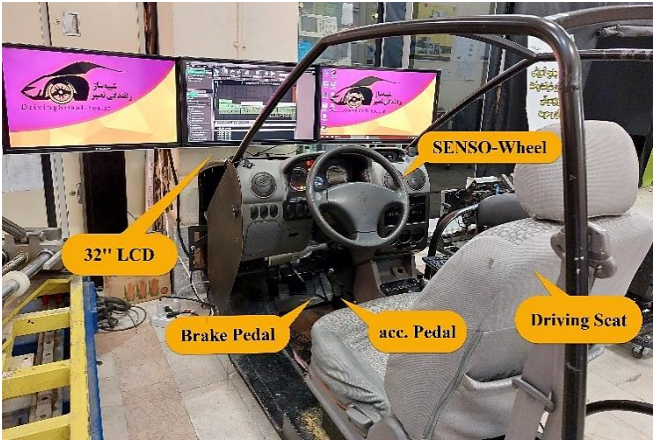

Figure 3. The used simulator in this study

*B. Data Collection*

To collect data, four expert drivers were asked to perform a slalom maneuver with a driving simulator located in Nasir Driving Simulator Lab at the K.N. Toosi University of Technology. As shown in Fig. 1, the slalom maneuver includes three sets of cones/obstacles that lead to three lane-changing and four longitudinal movements.

The four expert drivers carried out and repeated the maneuver 573 times with non-fixed speeds ranging from 15 to 60 kilometers. The data of 500 maneuvers were randomly selected for training set and the rest of the maneuvers for test set. The simulator constantly sampled and saved the driving data with a rate of 30 hertz per second. The driving data includes various information like speed, steering wheel angle, car head angle, x, y, tire angle, car horn, car gear, etc. Also, it's worth mentioning that the simulator software was developed using OpenGL and Python scripts. Fig. 3 shows the simulator located in Nasir Driving Simulator Lab.

*C. Feature Extraction and the CNN model*

In this project, after carefully investigating various features and machine learning models, we came to conclude that the performance of 2D CNN models with seven input features outperforms other models such as 1D convolutional neural network, LSTM, GRU, shallow MLP, KNN, and SVR. Detailed explanations on the extracted features and the structure of the model will be presented next.

As we know, the input to a CNN model must be a 2D vector. On the other hand, we extracted only seven features at each time step of driving that is actually a 1D vector. To convert these 1D vectors to 2D vectors, we constructed 5x7 matrices. We put the seven features with five different permutations in each row.

As mentioned, seven features were extracted. The first is about the state of the car's lateral motion. This feature has three constant values, which are 1, 2, and 3, that express turning left, turning right, and no turn (when moving in parallel of cone sets), respectively. We designed this feature inspired by drivers' behavior as a driver observes the vacant spaces ahead and decides to turn right or left. Also, when a driver moves parallel to the obstacle, the driver decides to move straight and not turning right or left. The driver senses the circumstance by her/his eyes while in our system, this feature is obtained by the simulator at each time step. Fig. 4 demonstrates this feature.

The second and third features are about lateral and longitudinal distances between the car head and the start/end of a cone set. When the car is heading a cone set, the distances are computed with regard to the start of the cone set. But, when the car is moving parallel to a cone set, the distances are calculated with regard to the end of the cone set. Fig. 5 demonstrates these features better. It's worth noting that the lateral distance is simply computed by subtracting $y_{car}$ from $y_{cone}$ and could be positive or negative (Eq. 1), whereas the longitudinal distance is obtained by the inverse of $1 + (x_{cone} - x_{car})$, which is observable in Eq. 2. Since $x_{cone}$ is always bigger than $x_{car}$ in the map of the simulator, the output of the subtraction is always positive. Consequently, the value of Eq. 2 is always between zero and one. When the car nears the start/end of an obstacle, the value of Eq. 2 nears one. Therefore, this feature becomes more meaningful.

$$f2 = y_{cone} - y_{car} \quad (1)$$

$$f3 = \frac{1}{1+(x_{cone}-x_{car})} \quad (2)$$

The fourth feature is the speed of the car at each time step that is between 15 km/h and 60 km/h. The fifth feature is car head angle relative to $y$-axis. The sixth and seventh features are the time derivative of the car head angle change and the rate of steering wheel angle change, respectively. We did not employ the previous steps of the steering wheel angle as input features because the output of the CNN model is actually the steering wheel angle. Therefore, previous steps of steering wheel angle convert our problem into a time series problem. During the project, whenever we employed the previous steps of steering wheel angle as input features, the model acted like a predictive model and couldn't be able to determine whether the steering wheel angle was appropriate or not. In other words, the model was predicting the trend of the steering wheel angle and didn't care about performing the maneuver correctly. Therefore, we didn't consider the previous steps of steering wheel angle for the input of the CNN model.

As mentioned before, we witnessed that CNN models work better for this specific task. It's highly likely that the CNN model can take the correlation between the features into consideration. The CNN model has been generally made up of three convolutional layers and three dense layers. All layers except the output layer exploit Exponential Linear Unit (ELU) function as activation function. The output layer is actually responsible for computing the steering wheel angle and employs a linear function as activation function. The three convolutional layers have 32, 64, and 128 kernels, respectively, with the same size of 2 by 2. The structure of the CNN model is shown in the Fig. 6.

Also, Adam optimization algorithm was used as the optimizer to train and update the parameters of the model. Moreover, it's worth noting that the adaptive learning rates led to better result when training the network. Therefore, we chose an adaptive learning rate with the initial value of 0.001 for Adam algorithm. The model was trained for 18 epochs with a batch size of 4. The Mean square error (MSE) was used to evaluate the performance of the model during the training process. The training procedure of the model is demonstrated in Fig. 7. After training the model, we evaluated the network's performance in terms of MSE and R2-squared criteria with training and test sets. R2-squared criterion is between zero and one; zero means no correlation and one shows complete correlation between the output of the network and the desired values. From Table 1, we can observe that the model managed to mimic the drivers' behavior with R2-squared of 0.83 for test set.

Table 1. Evaluation of the CNN model in terms of MSE and R2-squared criteria

| Metrics / Data sets | MSE | R2-squared |
|---|---|---|
| Training set | $3.76 \times 10^{-4}$ | 0.87 |
| Test set | $3.78 \times 10^{-4}$ | 0.83 |

*D. Control Unit*

In the previous part, we mentioned how the CNN model can find the patterns behind the scene of an expert driver's maneuvering. On the other hand, it can also imitate the general trend during the driving behavior in the maneuver. In this section, we're going to focus on implementing the designed model on the real steering wheel. To accomplish this part, we combined the Neural Network model and PD control-based rule:

$$\theta_d = F_{NN}(features) \quad (3)$$

$$e_\theta = \theta_a - \theta_d \quad (4)$$

$$\tau_G = P_g e_\theta + D_g \dot{e}_\theta \quad (5)$$

In the above equation, $\theta_a$ is the actual steering wheel angle, and $\theta_d$ is the desire steering wheel angle produced by the Neural Network function which has been demonstrated by $F_{NN}$. Using these gains ($P_g$, $D_g$) alongside the PD control-based rule on each time step, $\theta_a$ converges to $\theta_d$. Converging $\theta_a$ and $\theta_d$ will produce the proper torque to stimulate the steering wheel toward the desired angle in each time step. As a result, the vehicle can perform the maneuver autonomously. More information about how $P_g$ and $D_g$ will be presented in the future works.

III. RESULTS

In this section, we investigate the performance of the developed system. To evaluate our method, we executed the CNN model and the controller on the simulator in real-time. We observed that our method was able to steer the car perfectly and perform the slalom maneuver without any collision.

Fig. 8 depicts and compares the steering wheel angle taken by an expert driver and our method. In order to make a fair comparison, the experiment shown in Fig. 8 has been performed at the same initial location and based on the same speeds at each time step. According to Fig. 8, we can see that our method sometimes takes smoother angles. For example, by taking a meticulous look at the time interval between 5 and 10 seconds, we notice that our method takes a smoother steering wheel angle than the expert driver.

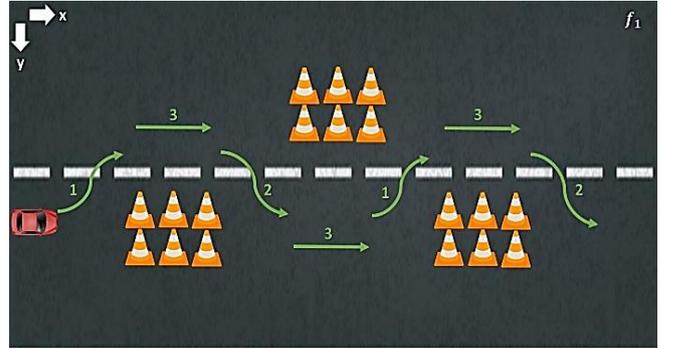

Figure 4. The graphical view of the first feature

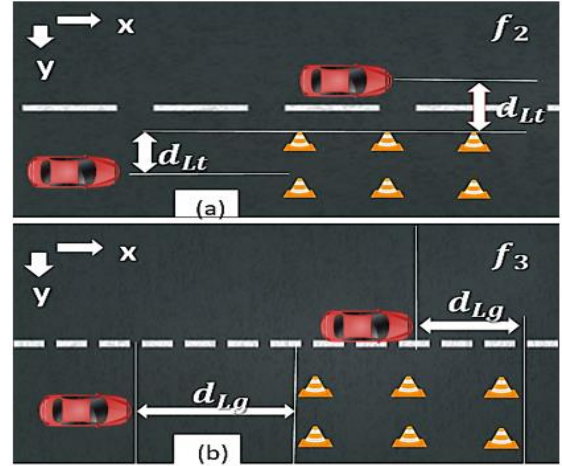

Figure 5. The first and the second features (lateral and longitudinal distances. a) Lateral distance. b) Longitudinal distance. $D_{Lt}$ and $D_{Lg}$ are lateral and longitudinal distances, respectively.

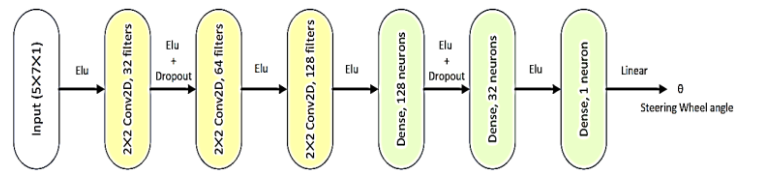

Figure 6. The structure of the CNN model

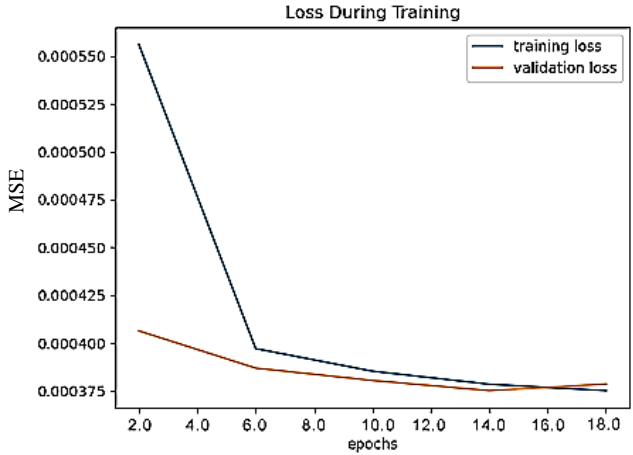

Figure 7. The CNN model's loss during training

Moreover, we can observe that at each positive or negative peak in Fig. 8, our system adopts a smaller angle compared to what the expert driver chooses, leading to a smoother obstacle avoidance. Also, the path and the coordination of the car at each time step for the expert driver and our system is shown in Fig. 9. In Fig. 9, we can see that our system crossed the cone sets smoothly and without any collision.

Fig. 10 shows the paths taken by the car in 17 trials. To perform the trials, the designed slalom maneuver was repeated 17 times at different longitudinal speeds. Our algorithm steered the car through the cone sets (red rectangles in Fig. 10) without any collision. It's worth mentioning that during the trials, our system only controlled the steering wheel and did not have any control over the accelerator pedal. This pedal was pressed at the discrete of the human driver. Therefore, as the car speed was not constant and was constantly changed by the driver, it was a big challenge for the algorithm to adapt itself and steer the car properly. The curves of speed and steering wheel angle for the mentioned 17 trials are illustrated in Fig. 11 and Fig. 12.

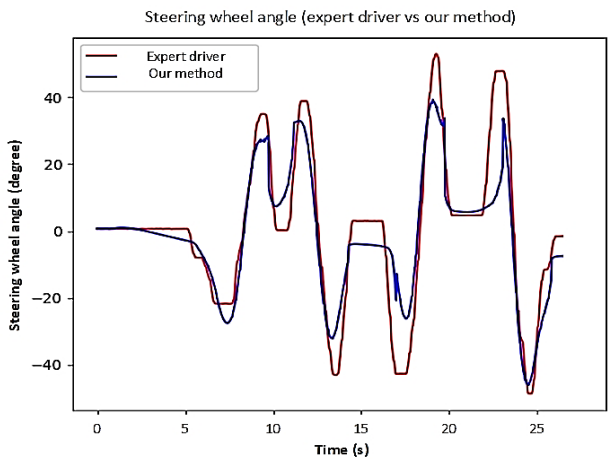

Figure 8. Steering wheel angle (our system vs. the expert driver)

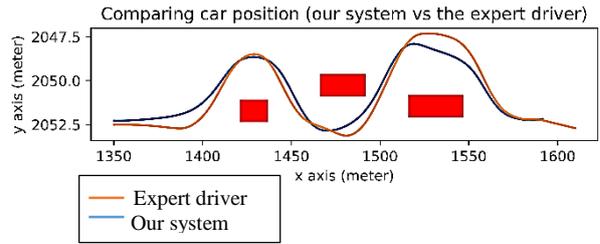

Figure 9. Position of the car (our system vs. the expert driver). Red rectangles are the cone sets.

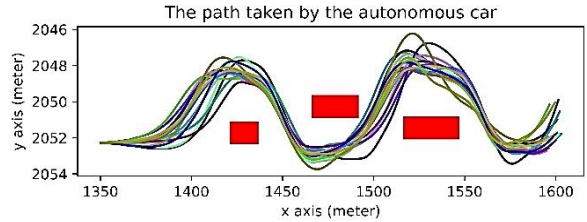

Figure 10. The paths taken by our system for 17 trials. The red rectangles are the cone sets.

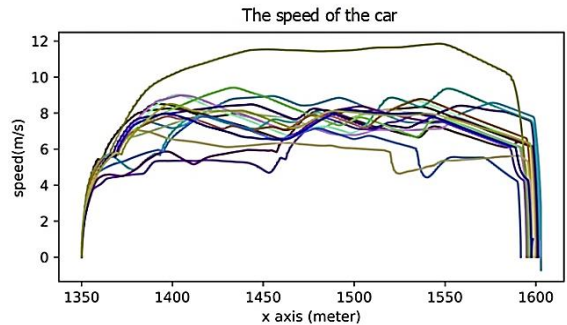

Figure 11. The speed applied to the car by the driver for 17 trials.

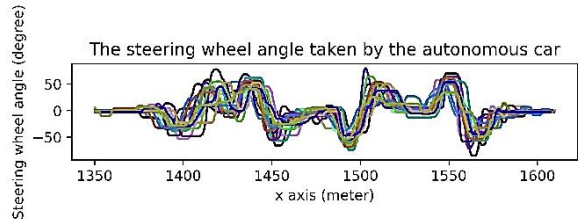

Figure 12. The steering wheel angle taken by our system for 17 trials.

## IV. DISCUSSION

As said before, there exist two main attitudes for developing a lane-changing and obstacle avoidance system -- path trajectory approaches and RL-based methods. However, the path taken by these approaches is not as smooth as the path taken by an expert driver. On the other hand, although this paper is about an automatic lane-changing and obstacle avoidance system, our main goal is to develop a haptic driving training system. Therefore, it's very important for us that the trainer system acts like an expert driver when performing a lane-changing maneuver. Hence, we decided to gather a large dataset from four expert drivers and train a

CNN model with the data to mimic the drivers' behavior and approximate the appropriate steering wheel angle.

The initial results are promising by examining the experimental results of the developed system on the simulator. The developed system was able to perform a complex slalom maneuver which constituted of three sets of cones.

Finally, we can conclude that, contrary to the common approaches like path trajectory and RL-based approaches, which have a specific and certain path or policy, training supervised algorithms could be an alternative method. However, this attitude has its merits and demerits. Mimicking the expert driver's behavior could be counted as a merit for this approach. Also, these methods could perfectly outperform the other mentioned approaches if the designer of the system extracts more meaningful features and also manipulates the cost function of the model, as is common in physics-informed neural works [12]. Regarding demerits, it should be said that gathering a dataset for each scenario could be time-consuming.

## V. Conclusion

In this paper, a new method was developed to perform slalom maneuvers by mimicking expert drivers' behavior. Seven meaningful features were extracted from the dataset gathered from four drivers. The features were converted to 2D vectors and then employed as the input of a CNN model, which is responsible for approximating the appropriate steering wheel angle. Then, the steering wheel angle computed by the CNN model was sent to the control unit, which consists of an adaptive PD controller. The control unit applies proper torque to the steering wheel of the driving simulator. Finally, the car performs the slalom maneuver and crosses the obstacle without any collision.

We observed that the designed system functions well and is able to perform the maneuver smoothly. Also, by comparing the curves of steering wheel angle taken by an expert driver and our method, we observed that not only the performance of the developed method is close to the expert driver (R2-squared of 0.83 based on Table 1), but also our method crosses the obstacles more safely and smoothly. In the future, we will develop a haptic driving training simulator that is able to teach novice drivers how to avoid obstacles.